\documentclass{article}
\usepackage{spconf,amsmath,graphicx,hyperref}

\usepackage{booktabs}
\usepackage{multirow}   
\usepackage{booktabs}   
\usepackage[table]{xcolor}  
\usepackage{subcaption}
\usepackage{graphicx} 
\usepackage{spconf,amsmath,graphicx,hyperref}
\usepackage{tabularx}
\usepackage{ragged2e}
\usepackage{graphics}
\usepackage{booktabs}
\usepackage{enumitem}
\usepackage{caption}  
\usepackage{float}    

\title{MAPEX: A Multi-Agent Pipeline for Keyphrase Extraction}
\name{Liting Zhang \qquad Shiwan Zhao \qquad Aobo Kong \qquad Qicheng Li$^\dagger$}

\address{TMCC, College of Computer Science, Nankai University, Tianjin, China\\
Email: 2120250707@mail.nankai.edu.cn}
\vspace{-2mm}
\begin{document}
%
\maketitle
\let\thefootnote\relax\footnotetext{\raggedright $\dagger$ Corresponding author.}

%
%
\vspace{-2mm} 


\begin{abstract}




Keyphrase extraction is a fundamental task in natural language processing. However, existing unsupervised prompt-based methods for Large Language Models (LLMs) often rely on single-stage inference pipelines with uniform prompting, regardless of document length or LLM backbone. Such one-size-fits-all designs hinder the full exploitation of LLMs' reasoning and generation capabilities, especially given the complexity of keyphrase extraction across diverse scenarios. To address these challenges, we propose MAPEX, the first framework that introduces multi-agent collaboration into keyphrase extraction. MAPEX coordinates LLM-based agents through modules for expert recruitment, candidate extraction, topic guidance, knowledge augmentation, and post-processing. A dual-path strategy dynamically adapts to document length: knowledge-driven extraction for short texts and topic-guided extraction for long texts. Extensive experiments on six benchmark datasets across three different LLMs demonstrate its strong generalization and universality, outperforming the state-of-the-art unsupervised method by 2.44\% and standard LLM baselines by 4.01\% in F1@5 on average. Code is available at \url{https://github.com/NKU-LITI/MAPEX}. 








\end{abstract}

\vspace{-2mm}
\section{INTRODUCTION}
\label{sec:intro}

Keyphrase extraction is a fundamental task in natural language processing, focusing on automatically identifying phrases that concisely represent the core content of a document. This task serves as a critical building block for downstream applications including information retrieval, text summarization and recommendation systems. 
It can be broadly divided into supervised and unsupervised approaches. However, supervised methods require large-scale labeled training data and generalize poorly to new domains, whereas unsupervised approaches are more widely adopted in practice.


Traditional unsupervised keyphrase extraction methods can be categorized into three types\cite{category}. Statistical methods~\cite{TFIDF,yake} exploit features of candidates such as frequency and position, while graph-based methods~\cite{mihalcea2004textrank,singlerank,bougouin2013topicrank,florescu2017positionrank, MultipartiteRank} model the text as a network and apply centrality measures. 
More recently, embedding-based methods~\cite{embedrank,SIFRank,mderank} leverage pre-trained language models (PLMs) embeddings to estimate semantic relevance, achieving competitive performance. PromptRank~\cite{promptrank} introduces a prompt-based paradigm for PLMs which ranks candidates based on their generation probability within a specific prompt template. With the rise of large language models (LLMs), these prompt-based approaches have been further applied to and enhanced by LLMs. Later work has explored diverse prompting strategies such as the combination of vanilla and candidate-based prompting~\cite{Hybrid}, and proposed few-shot extensions that integrate fine-grained feedback into in-context learning~\cite{huang2023enhancingincontextlearninganswer}. 
However, existing prompt-based methods for LLMs typically rely on single-stage inference pipelines that apply a uniform prompting strategy regardless of document length~\cite{song2024preliminaryempiricalstudypromptbased}, limiting their generalization across different LLM backbones.




Concurrently, recent studies have shown that incorporating agent-based techniques with large language models, such as multi-role expert agents~\cite{roleplay1,roleplay2} or reasoning–action integration frameworks~\cite{yao2023react}, has yielded promising results in complex tasks. 
Keyphrase extraction is also inherently multi-stage, typically involving candidate extraction and ranking, which makes it a natural fit for agent-based designs. 
Nevertheless, the application of such mechanisms to this task remains underexplored, and a generalizable pipeline has not yet been developed.




To address these challenges, we propose MAPEX, a multi-agent pipeline framework for keyphrase extraction in zero-shot setting. MAPEX coordinates three agents through a pipeline of multiple modules, including expert recruitment, candidate extraction, topic identification, knowledge augmentation, candidate re-ranking, and post-processing. It also adopts tailored strategies for documents of different lengths. The main contributions of this paper are as follows:

\begin{enumerate}[leftmargin=*]
    \item We first integrate multi-agent collaboration into keyphrase extraction task, leveraging role-based prompting, candidate extraction, topic guidance, knowledge augmentation, and post-processing to enhance LLMs' extraction quality.
    \item We propose MAPEX, a dual-pipeline framework that dynamically adapts to document length by incorporating knowledge-driven extraction for short texts and topic-guided extraction for long texts, effectively addressing the challenges of semantic dilution and context window limitations in LLMs.
    \item We conduct comprehensive experiments on six benchmark datasets across three different LLMs, demonstrating the strong generalizability and universality of MAPEX. It consistently outperforms state-of-the-art (SOTA) unsupervised method and standard LLM baselines, while also exhibiting robustness in complex contexts.
\end{enumerate}

\vspace{-2mm}
\section{METHOD}
\label{method}



\begin{figure}[htp]
\begin{minipage}[b]{1.0\linewidth}
  \centering
  \centerline{\includegraphics[width=8.5cm]{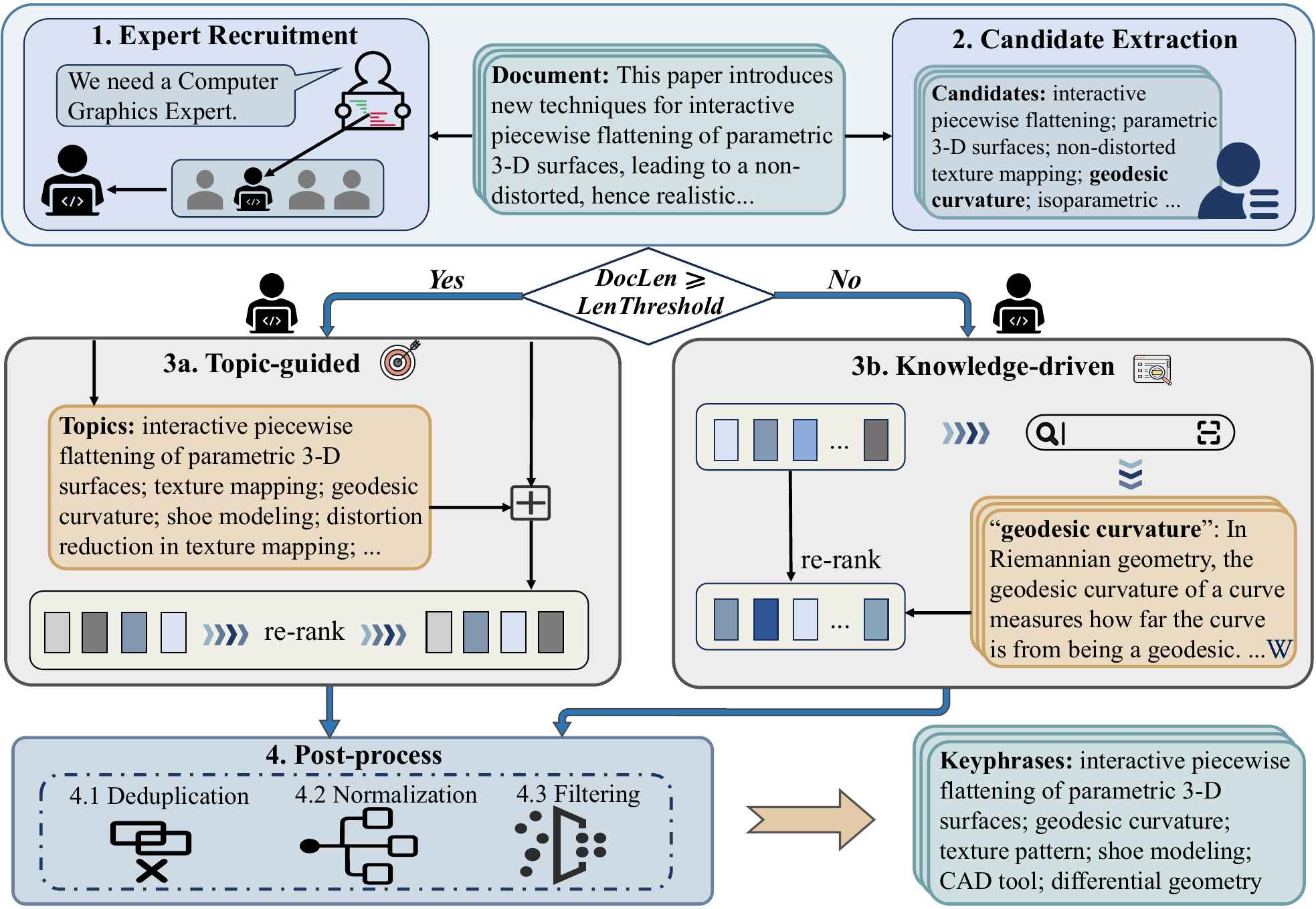}}
\end{minipage}
\caption{MAPEX framework. The three main agents are: Expert Recruiter, Candidate Extractor, and Domain Expert.}
\label{fig:method}
\end{figure}

In this section, we introduce a multi-agent pipeline for keyphrase extraction. As shown in Fig.~\ref{fig:method}, the framework involves three agents, an Expert Recruiter, a Candidate Extractor and a Domain Expert, operating under a dual-path strategy conditioned on document length. 


\vspace{-2mm} 
\subsection{Expert Recruitment}
To enhance the specificity and effectiveness of keyphrase extraction, we introduce the Expert Recruiter agent. This agent is responsible for dynamically defining the role of the Domain Expert agent, who conducts the extraction from a specialized domain perspective. 
Specifically, given a document, this Expert Recruiter first analyzes its domain characteristics and then assigns an appropriate expert role (e.g., computer graphics expert, mathematician, or software engineer) along with a rationale for this assignment. The recruited Domain Expert subsequently participates in the dual-path pipeline strategy, where it provides crucial domain-specific guidance for topic identification and candidate re-ranking.

\begin{figure}[t]
\begin{minipage}[b]{1.0\linewidth}
  \centering
  \includegraphics[width=8.5cm]{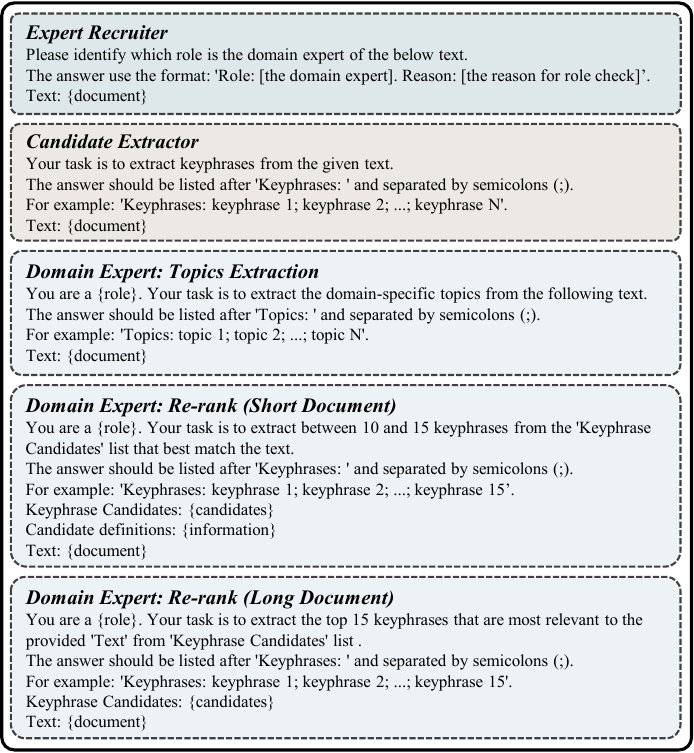}
\end{minipage}
\caption{Prompts used in the pipeline.}
\label{fig:prompt}
\end{figure}

The detailed prompt designs for all three agents are illustrated in Fig.~\ref{fig:prompt}. 

\vspace{-2mm}
\subsection{Candidate Extraction}

After establishing the role of the Domain Expert, the next step is to obtain a broad set of potential keyphrase candidates from the document. Since relying solely on the expert agent may overlook diverse lexical variations~\cite{mind}, we introduce a common candidate extraction process to mitigate this risk. 
The Candidates Extractor is responsible for generating an initial pool of keyphrase candidates directly from the document, without assigning any specific role. 

\vspace{-2mm}
\subsection{Dual-Path Pipeline Strategy}

To handle documents of varying lengths, our framework introduces a length threshold $\ell$ to differentiate short and long documents. This dual-path design is motivated by the empirical finding that external knowledge infusion provides substantial gains for short documents, while its relative advantage diminishes on longer texts. We hypothesize that this is due to semantic dilution and the limited context window of LLMs, which hinder effective knowledge grounding. Accordingly, the Domain Expert processes short documents by routing them through a knowledge-driven extraction path, while long documents are handled by a topic-guided pipeline to distill core themes. 

For long documents ($|d_i| \ge \ell$), the topic-guided path is employed. The Domain Expert generates a list of salient topics representing the document's core themes, which are combined with candidates extracted by the Candidate Extractor in the previous step. Subsequently, it re-ranks and filters this enriched candidate list to produce a preliminary keyphrase list. 

For short documents ($|d_i| < \ell$), the knowledge-driven path is activated. To enhance the semantic understanding of candidate keyphrases, the Domain Expert performs an external knowledge retrieval action from sources such as Wikipedia to enrich candidates with definitions and contextual information. For each candidate keyphrase $c_{ij}$, the Domain Expert invokes $\mathrm{WikiQuery}(c_{ij})$ to retrieve a concise summary. The aggregated knowledge from all candidates forms a dictionary: 
\begin{equation*}
W_i = \bigcup_{c \in C_i} \mathrm{WikiQuery}(c).
\end{equation*}
Similarly, based on this augmented information, the Domain Expert re-ranks the candidates extracted in the previous step and outputs a preliminary keyphrase list. 


\vspace{-2mm}
\subsection{Post-process}

After the Domain Expert produces a preliminary keyphrase list, MAPEX applies a final post-processing step to mitigate hallucinations (i.e., spurious or non-grounded keyphrases) and reduce noise in the outputs. This involves three consecutive steps: (1) removing redundant phrases, (2) normalizing abbreviations and their full forms, and (3) filtering out phrases absent in the original text. This ensures the final keyphrases are concise, accurate and contextually valid. 

\vspace{-2mm}
\section{EXPERIMENTS}
\label{exper}


\vspace{-2mm}
\subsection{Datasets}
We empirically evaluate MAPEX on six widely used benchmark datasets: Inspec~\cite{inspec}, SemEval-2010~\cite{semeval2010}, SemEval-2017~\cite{semeval2017}, DUC2001~\cite{duc2001}, NUS~\cite{nus}, and Krapivin~\cite{krapivin}. 
The statistics information of datasets are summarized in Table~\ref{tab:dataset_stats}.


\vspace{-2mm} 
\subsection{Baselines}


We compare MAPEX against representative unsupervised baselines. The selected baselines are: 
(1) Embedding-based Methods: SIFRank~\cite{SIFRank} uses SIF-weighted embeddings and PageRank to rank candidates. MDERank~\cite{mderank} uses masked document embeddings to better capture keyphrase importance. 
(2) PLM-based Prompted Method: PromptRank~\cite{promptrank} uses a BART encoder–decoder and ranks candidates by the decoder’s generation probability under prompts. 
(3) LLM-based Prompted Methods: Base denotes a naive prompting baseline, while Hybrid~\cite{Hybrid} combines vanilla and candidate-based prompting to improve zero-shot performance. 

\vspace{-2mm} 
\subsection{Experimental Setup}




All experiments are conducted on two NVIDIA GeForce RTX 3090 24GB. We evaluate MAPEX on three open-source LLMs: Mistral-7B-Instruct-v0.3, Qwen2-7B-Instruct, and Qwen2.5-7B-Instruct. To maximize inference efficiency, we utilize the vLLM engine and Hugging Face Transformers for batch processing. To ensure reproducibility, we use greedy decoding strategy by setting \texttt{do\_sample=False} in Hugging Face and \texttt{temperature=0} in vLLM. For the MAPEX pipeline, we set the LengthThreshold ($\ell$) to 512 tokens. Following the source code released by PromptRank~\cite{promptrank}, we adopt the same data preprocessing and evaluation protocol as Hybrid~\cite{Hybrid}, using F1-score as the evaluation metric.


\begin{table}[t]
    \centering
    \caption{Statistics of the benchmark datasets.
    \#Doc: number of documents, AvgTok: average tokens per document, \#Gold: number of gold keyphrases in each dataset.}
    \small
    \begin{tabular}{l l r r r} 
        \toprule
        Dataset & Domain & \#Doc & AvgTok & \#Gold \\
        \midrule
        Inspec & Scientific & 500 & 166 & 4,912 \\
        SemEval2017 & Scientific & 493 & 229 & 8,387 \\
        SemEval2010 & Scientific & 100 & 253 & 1,506 \\
        DUC2001 & News & 307 & 1,063 & 2,479 \\
        NUS & Scientific & 211 & 10,685 & 2,453 \\
        Krapivin & Scientific & 460 & 11,423 & 2,641 \\
        \bottomrule
    \end{tabular}
    \label{tab:dataset_stats}
\end{table}

\vspace{-2mm}
\subsection{Overall Results}


As summarized in Table~\ref{overall}, MAPEX consistently outperforms both traditional unsupervised methods and standard LLM baselines across six benchmark datasets and multiple LLM backbones. 
When compared against traditional unsupervised methods, MAPEX achieves the highest average F1@5 score of 24.30\% when using Qwen2.5-7B-Instruct, outperforming the previous state-of-the-art method PromptRank (22.81\%) by 2.44\%. Notably, it achieves significant gains on long documents, with a +5.22\% absolute improvement on NUS. 

Compared to the standard LLM baseline (Base), MAPEX shows consistent improvements across all model and dataset configurations. In particular, MAPEX brings substantial performance boosts to all underlying LLMs, improving the average F1@5 score of Mistral-7B by 4.01\% (from 18.23\% to 22.24\%), Qwen2-7B by 1.45\%, and Qwen2.5-7B by 1.14\%. These improvements are consistently observed across F1@10 and F1@15 metrics, underscoring the robustness and generalizability of MAPEX across diverse model architectures. 



\begin{table}[htbp]
\centering
\caption{Performance of MAPEX and baselines on F1@K across datasets.}
\label{overall}
\setlength{\tabcolsep}{2pt} 
{\scriptsize 
\resizebox{\columnwidth}{!}{%
\begin{tabular}{clccccccc} 
\toprule
\multirow{2}{*}{$F_{1}@K$}    & \multirow{2}{*}{Method} & \multicolumn{6}{c}{Dataset}                                    & \multirow{2}{*}{AVG}  \\ 
\cmidrule{3-8}
                     &                          & Inspec & Sem2017 & Sem2010 & DUC2001 & NUS & Krapivin   &                       \\ 
\midrule
\multirow{10}{*}{5}  & SIFRank (ELMo)           & 29.38  & 22.38       & 11.16       & 24.30   & 3.01     & 1.62  & 15.31                 \\
                     & MDERank (BERT)           & 26.17  & 22.81       & 12.95       & 13.05   & 15.24    & 11.78 & 17.00                 \\
                     & PromptRank (T5)          & 31.73  & \textbf{27.14}   & \textbf{17.24}  & 27.39   & 17.24   & 16.11 & 22.81     \\
                     & Hybrid (Mistral-7B)      & 31.18  & 19.41   & 12.06  & 23.16   & 15.13   & 15.42 & 19.39     \\
\cmidrule{2-9}
		            & Base (Mistral-7B)         & 30.17 & 18.23 & 11.76 & 23.01 & 12.91 & 13.28 & 18.23 \\
                    & MAPEX (Mistral-7B)        & 33.78 & 22.95 & 15.35 & \textbf{27.53} & 17.37 & 16.49 & 22.24 \\
\cmidrule{2-9}
                    & Base (Qwen2-7B)           & 32.72 & 19.48 & 12.95 & 21.77 & 19.65 & 19.92 & 21.08 \\
                    & MAPEX (Qwen2-7B)          & 33.13  & 21.06       & 13.95       & 24.83   & 21.39    & 20.84 & 22.53               \\
                    & Base (Qwen2.5-7B)         & 35.23 & 22.56 & 16.74 & 25.26 & 20.88 & 18.29 & 23.16 \\
                    & MAPEX (Qwen2.5-7B)        & \textbf{35.43}  & 24.14       & 16.44       & 25.81   & \textbf{22.46}    & \textbf{21.54} & \textbf{24.30}               \\

\midrule 
\midrule
\multirow{10}{*}{10} & SIFRank (ELMo)           & 39.12  & 32.60       & 16.03       & 27.60   & 5.34     & 2.52  & 20.54                 \\
                     & MDERank (BERT)           & 33.81  & 32.51       & 17.07       & 17.31   & 18.33    & 12.93 & 21.99                 \\
                     & PromptRank (T5)          & 37.88  & \textbf{37.76}   & \textbf{20.66}   & \textbf{31.59}   & 20.13   & 16.71 & 27.46  \\    
                     & Hybrid (Mistral-7B)      & 39.53  & 29.33   & 17.10  & 25.84   & 17.46   & 16.19 & 24.24     \\
\cmidrule{2-9}
                    & Base (Mistral-7B)         & 37.73 & 27.23 & 15.80 & 25.45 & 14.92 & 14.22 & 22.56 \\
                    & MAPEX (Mistral-7B)        & 41.54 & 33.53 & 19.28 & 27.90 & 19.94 & 16.36 & 26.42 \\
\cmidrule{2-9}      
                    & Base (Qwen2-7B)           & 40.70 & 29.87 & 17.27 & 24.18 & 21.43 & 19.59 & 25.51 \\
                    & MAPEX (Qwen2-7B)          & 41.37  & 31.51       & 18.37       & 26.58   & 23.57    & \textbf{21.31} & 27.12               \\
                    & Base (Qwen2.5-7B)         & \textbf{42.97} & 32.28 & 20.14 & 25.93 & 23.12 & 18.80 & 27.21 \\
                    & MAPEX (Qwen2.5-7B)        & 42.42  & 34.56       & 20.21       & 26.74   & \textbf{24.28}    & 21.03 & \textbf{28.21}               \\        
\midrule
\midrule
\multirow{10}{*}{15} & SIFRank (ELMo)           & 39.82  & 37.25       & 18.42       & 27.96   & 5.86     & 3.00  & 22.05                 \\
                     & MDERank (BERT)           & 36.17  & 37.18       & 20.09       & 19.13   & 17.95    & 12.58 & 23.85                 \\
                     & PromptRank (T5)          & 38.17  & \textbf{41.57}   & \textbf{21.35}  & 31.01   & 20.12  & 16.02 & 28.04  \\
                     & Hybrid (Mistral-7B)      & 40.69  & 33.40   & 17.75  & 24.43   & 17.16   & 15.10 & 24.76     \\
\cmidrule{2-9}
                     & Base (Mistral-7B)        & 38.72 & 32.89 & 16.97 & 24.72 & 15.43 & 13.71 & 23.74 \\
                     & MAPEX (Mistral-7B)       & 43.04 & 37.44 & 19.38 & \textbf{26.52} & 19.33 & 15.68 & 26.90 \\
\cmidrule{2-9}      
                    & Base (Qwen2-7B)           & 43.12 & 35.71 & 18.48 & 22.88 & 20.73 & 17.31 & 26.37 \\
                    & MAPEX (Qwen2-7B)          & \textbf{43.86}  & 37.77       & 19.49       & 25.26   & 23.41    & \textbf{20.69} & 28.41               \\
                    & Base (Qwen2.5-7B)         & 42.26 & 36.46 & 20.70 & 24.03 & 21.25 & 17.16 & 26.98 \\
                    & MAPEX (Qwen2.5-7B)        & 42.75  & 37.95       & 20.89       & 26.32   & \textbf{23.71}    & 19.86 & \textbf{28.58}               \\
\bottomrule
\end{tabular} } }
\end{table}


\vspace{-2mm} 

\subsection{Ablation Studies}


We conduct ablation studies on Mistral-7B-Instruct-v0.3 to evaluate the contributions of each component in MAPEX. Specifically, we analyse the efficacy of individual modules and empirically determine the length threshold within the performance crossover region of the two pipelines.

We perform a component-wise ablation study, starting from a Base setup of standard LLM inference without incorporating any specialized modules, corresponding solely to the Candidate Extraction step. 
We then iteratively construct the MAPEX pipeline by introducing modules step-by-step: first the Expert Role module, then the Topic or the Knowledge branch, and finally the Post-process module. The results are summarized in Table~\ref{component_ablation}. 
Introducing the Expert Role module yields consistent improvements on long-text datasets (e.g., NUS and Krapivin), though gains on short-text datasets are less stable. 
The Topic and Knowledge branches provide substantial improvements, especially on long-text datasets, highlighting their effectiveness in mitigating semantic dilution and enhancing candidate quality. Finally, the Post-process module further refines the predictions, enabling the full MAPEX pipeline to consistently achieve strong performance across datasets. Overall, these results demonstrate that each core component contributes positively. 

\begin{table}[htb]
\centering
\caption{Component ablation study results on Mistral-7b.}
\label{component_ablation}
\setlength{\tabcolsep}{2pt} 
{\scriptsize 
\resizebox{\columnwidth}{!}{%
\begin{tabular}{clccccccc} 
\toprule
\multirow{2}{*}{$F_{1}@K$}    & \multirow{2}{*}{Method} & \multicolumn{6}{c}{Dataset}                                    & \multirow{2}{*}{AVG}  \\ 
\cmidrule{3-8}
                     &                   & Inspec & Sem2017 & Sem2010 & DUC2001 & NUS & Krapivin   &     \\ 
\midrule
\multirow{5}{*}{5}  & Base               & 30.17 & 18.23 & 11.76 & 23.01 & 12.91 & 13.28 & 18.23 \\
                    & Expert Role+              & 29.74 & 18.12 & 11.46 & 20.87 & 13.20 & 13.56 & 17.82 \\
\cmidrule{2-9}
                    & Topic+            & 32.28 & 19.69 & 14.05 & 25.35 & 16.27 & 15.47 & 20.52 \\
                    & Post-process+      & 33.71 & 22.81 & 15.32 & 27.53 & 17.37 & 16.49 & 22.20 \\
\cmidrule{2-9}                 
                    & Knowledge+         & 32.54 & 20.50 & 14.55 & 24.00 & 15.87 & 14.35 & 20.30 \\
                    & Post-process+      & 33.78 & 22.95 & 15.35 & 26.76 & 16.79 & 16.04 & 21.94 \\
\midrule
\multirow{5}{*}{10} & Base               & 37.73 & 27.23 & 15.80 & 25.45 & 14.92 & 14.22 & 22.56 \\
                    & Expert Role+              & 37.51 & 27.29 & 15.40 & 22.53 & 14.96 & 14.58 & 22.05 \\
\cmidrule{2-9}
                    & Topic+            & 37.84 & 28.84 & 17.82 & 25.34 & 18.15 & 15.54 & 23.92 \\
                    & Post-process+      & 39.62 & 33.01 & 18.70 & 27.90 & 19.94 & 16.36 & 25.92 \\
\cmidrule{2-9}                 
                    & Knowledge+         & 40.55 & 30.58 & 17.91 & 25.24 & 17.23 & 15.07 & 24.43 \\
                    & Post-process+      & 41.54 & 33.53 & 19.28 & 26.96 & 18.72 & 15.81 & 25.97 \\
\midrule
\multirow{5}{*}{15} & Base               & 38.72 & 32.89 & 16.97 & 24.72 & 15.43 & 13.71 & 23.74 \\
                    & Expert Role+              & 39.08 & 33.01 & 17.40 & 21.95 & 15.39 & 13.78 & 23.43 \\
\cmidrule{2-9}
                    & Topic+            & 38.18 & 31.70 & 17.79 & 23.40 & 17.41 & 14.52 & 23.83 \\
                    & Post-process+      & 40.41 & 36.32 & 18.85 & 26.52 & 19.33 & 15.68 & 26.18 \\
\cmidrule{2-9}                 
                    & Knowledge+         & 41.77 & 34.31 & 18.53 & 23.27 & 16.47 & 13.84 & 24.70 \\
                    & Post-process+      & 43.04 & 37.44 & 19.38 & 24.59 & 18.07 & 14.57 & 26.18 \\
\bottomrule
\end{tabular} } }
\end{table}

\begin{figure}[tb]
\begin{minipage}[b]{0.48\linewidth}
  \centering
  \centerline{\includegraphics[width=4.0cm]{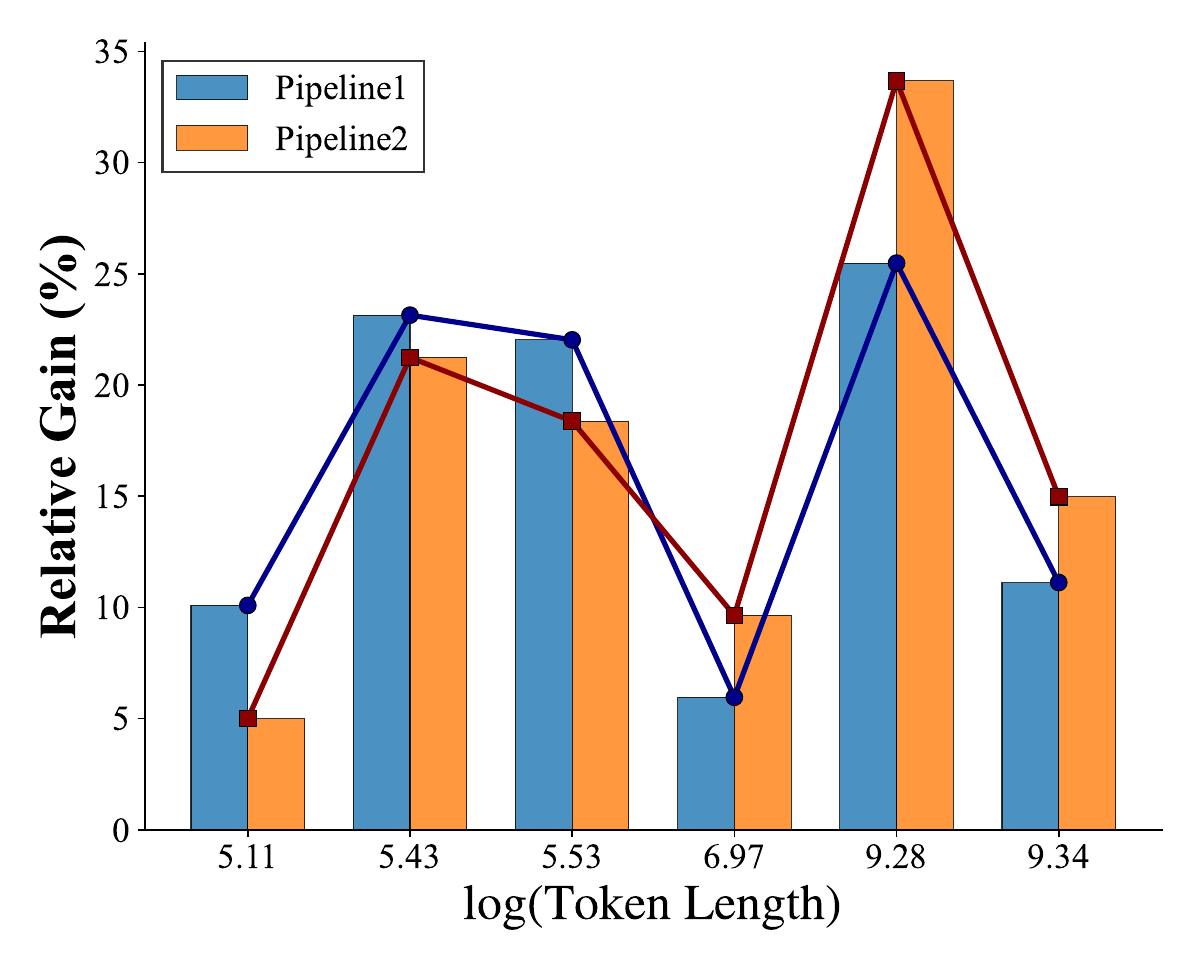}}
  \centerline{(a) Relative gain over Base}\medskip
\end{minipage}
\hfill
\begin{minipage}[b]{0.48\linewidth}
  \centering
  \centerline{\includegraphics[width=4.0cm]{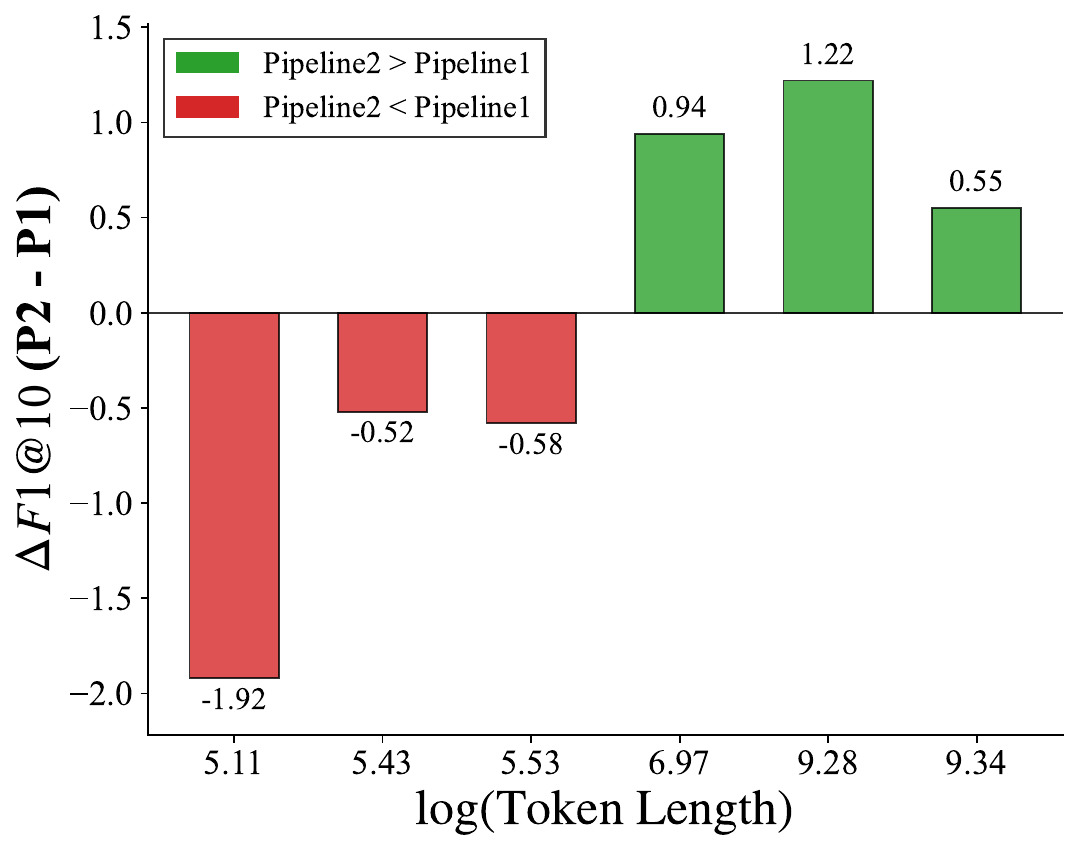}}
  \centerline{(b) Pipeline difference}\medskip
\end{minipage}
\caption{Length-based performance for routing threshold.}
\label{fig:lengththreshold}
\end{figure}

To determine the routing threshold based on input length, we group the samples into bins according to their token counts (the x-axis in Fig.~\ref{fig:lengththreshold} corresponds to the center of each bin in terms of $\ln(\text{length})$). For a given dataset $d$ and bin $b$, let $S_{d,b}$ denote the set of samples within that bin. We define
\begin{equation*}
F_i^{(d)}(b) = \mathrm{F1@10}(\text{pipeline } i; S_{d,b})
\end{equation*}
as the F1@10 score of pipeline $i$ evaluated on $S_{d,b}$. The absolute difference and relative improvement compared to the Base pipeline are computed as:
\begin{equation*}
\Delta_{i,\mathrm{Base}}^{(d)}(b)=F_i^{(d)}(b)-F_{\mathrm{Base}}^{(d)}(b),
\:
R_{i,\mathrm{Base}}^{(d)}(b)=\frac{\Delta_{i,\mathrm{Base}}^{(d)}(b)}{F_{\mathrm{Base}}^{(d)}(b)}.
\end{equation*}
Fig.~\ref{fig:lengththreshold} (a) shows, versus $\ln(\text{length})$, the averaged relative improvement $\frac{1}{D}\sum_{d} R_{i,\mathrm{Base}}^{(d)}(b)$ , while Fig.~\ref{fig:lengththreshold} (b) presents the averaged absolute difference $\frac{1}{D}\sum_{d}\Delta_{2,1}^{(d)}(b)$. A transition region is clearly observed for $\ln(\text{length}) \in [5.53, 6.97]$. We choose 512 tokens ($\ln(512) \approx 6.24$) as the routing threshold, which is close to this midpoint and consistent with the token truncation applied during data preprocessing. 

\section{CONCLUSION}
We present MAPEX, a multi-agent pipeline for keyphrase extraction, where LLM-based agents collaborate through specialized roles. 
The framework integrates expert recruitment, candidate extraction, topic guidance, knowledge augmentation, candidate re-ranking and post-process checks, and employs a dual-path strategy to handle documents of varying lengths: knowledge-driven extraction for short texts and topic-guided for long documents. 
Extensive evaluations on six benchmark datasets demonstrate that MAPEX consistently boosts performance and generalizes effectively across multiple LLMs, outperforming both state-of-the-art unsupervised methods and standard LLM baselines with robustness.


\vfill\pagebreak

\small
\bibliographystyle{IEEEbib}
\bibliography{arxiv,refs}

\end{document}